\documentclass{article}
\usepackage{spconf,amsmath,epsfig}
\usepackage[switch]{lineno}

\usepackage{multirow}
\usepackage{amsmath}
\usepackage{amssymb}
\usepackage{float} 
\usepackage{subfigure} 
\usepackage{booktabs}

\usepackage{paralist}

\let\OLDthebibliography\thebibliography
\renewcommand\thebibliography[1]{
  \OLDthebibliography{#1}
  \setlength{\parskip}{0pt}
  \setlength{\itemsep}{0pt plus 0.3ex}
}
\usepackage{bbding}
\usepackage{tikz}
\def\checkmark{\tikz\fill[scale=0.4](0,.35) -- (.25,0) -- (1,.7) -- (.25,.15) -- cycle;} 

\usepackage{colortbl}
\definecolor{maroon}{cmyk}{0,0.87,0.68,0.32}

\def\ucn{UCN}
\def\ucnA{Uncertainty-Based Network}

\def\miniimagenet{\textit{mini}ImageNet}
\def\tieredimagenet{\textit{tiered}ImageNet}

\floatstyle{ruled}
\newfloat{listing}{tb}{lst}{}
\floatname{listing}{Listing}

\begin{document}\sloppy

\def\x{{\mathbf x}}
\def\L{{\cal L}}

\title{Uncertainty-based Network for Few-shot Image Classification}
%

\name{Minglei Yuan, Qian Xu, Chunhao Cai, Yin-Dong Zheng, Tao Wang, Tong Lu$^{*}$\thanks{This work is supported by the Natural Science Foundation of China under Grant 61672273 and Grant 61832008, and Scientific Foundation of State Grid Corporation of China (Research on Ice-wind Disaster Feature Recognition and Prediction by Few-shot Machine Learning in Transmission Lines).} and Wenbin Li}
\address{National Key Laboratory for Novel Software Technology, Nanjing University \\
        \{mlyuan, chinyxu\}@smail.nju.edu.cn, chunhao.cai@smartmore.com, \\
        \{ydzheng0331, taowangzj\}@gmail.com, \{lutong, liwenbin\}@nju.edu.cn}

\maketitle

\begin{abstract}
The transductive inference is an effective technique in the few-shot learning task, where query sets update prototypes to improve themselves.
However, these methods optimize the model by considering only the classification scores of the query instances as confidence while ignoring the uncertainty of these classification scores.
In this paper, we propose a novel method called \ucnA{}, which models the uncertainty of classification results with the help of mutual information.
Specifically, we first data augment and classify the query instance and calculate the mutual information of these classification scores.
Then, mutual information is used as uncertainty to assign weights to classification scores, and the iterative update strategy based on classification scores and uncertainties assigns the optimal weights to query instances in prototype optimization.
Extensive results on four benchmarks show that \ucnA{}  achieves comparable performance in classification accuracy compared to state-of-the-art methods.

\end{abstract}
\section{Introduction}
\label{sec:introduction}


Few-shot image classification aims to learn knowledge with a few labeled samples quickly.
Recent few-shot image classification methods can be divided into two types.
The first type of method~\cite{protonet, matching, rn} is called inductive inference, where each unlabeled instance is predicted independently.
Although these methods perform well, they ignore the information of unlabeled query instances.
Thus, another type of method called transductive inference~\cite{propagatelabel, ent} is proposed to explore the information of the unlabeled instances.
Since transduction inference methods need to use pseudo-labels of unlabeled instances in the query set during the training phase, they often use few-shot learning models to obtain pseudo-labels and adopt classification scores as confidence.

\begin{figure}[t]
\centering
\includegraphics[width=0.9\linewidth]{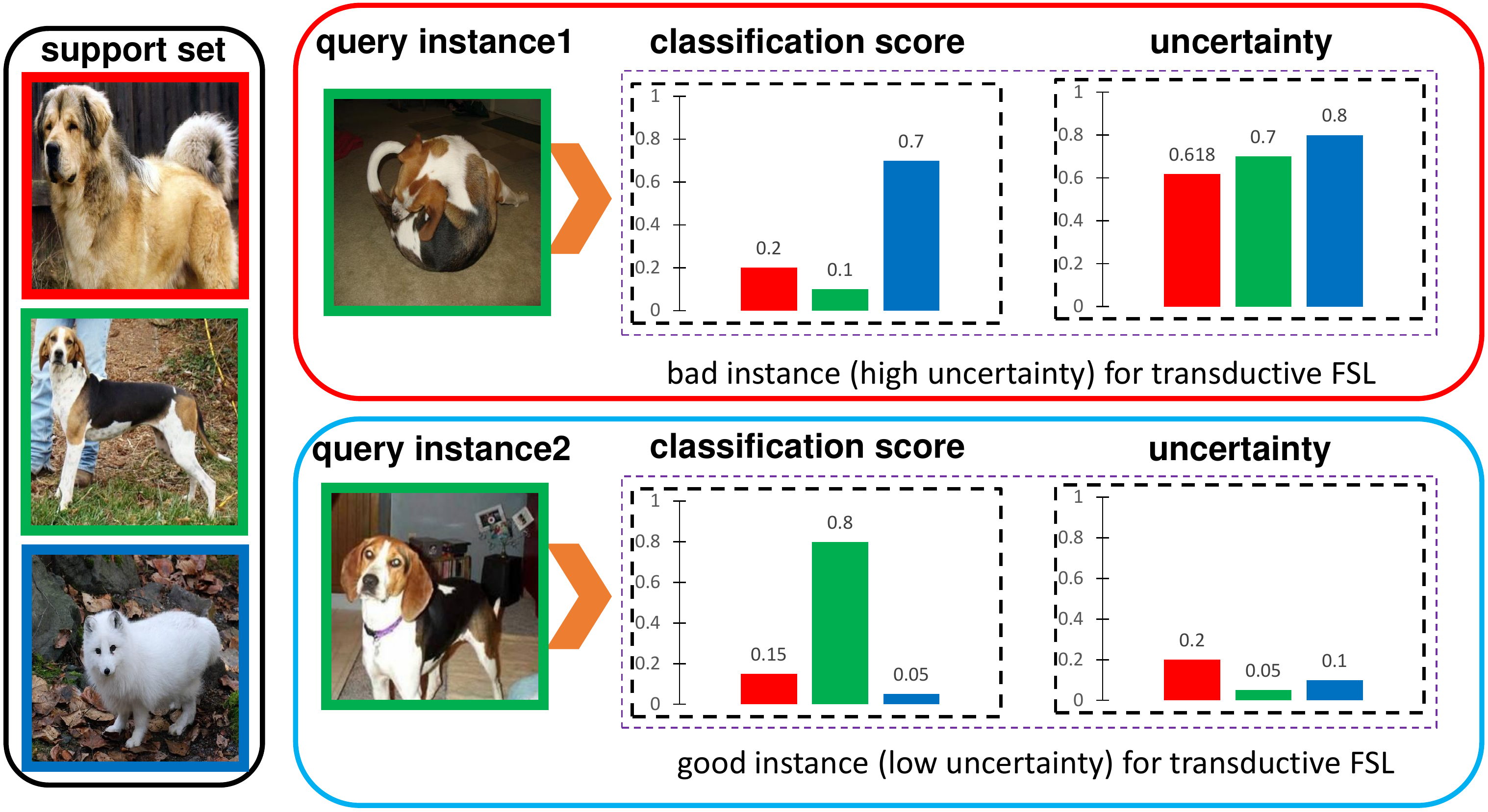}
\caption{\textbf{Examples of classification scores and uncertainty.} Two query instances in 3-way 1-shot transductive FSL, the smaller the uncertainty, the higher the confidence of the classification result.}
\vspace{-0.1in}
\label{fig:intr}
 \vspace{-0.1in}
\end{figure}

Although these transduction inference methods usually have better performance than inductive methods, due to the lack of sufficient data to enhance the generalization ability of the model, the models trained with few samples often obtain classification scores with high uncertainty.
This limitation can affect the effectiveness of the transductive FSL methods, which only use classification scores to refine prototypes.
Fig.~\ref{fig:intr} shows that a high classification score may have high uncertainty. If only classification results are considered, query instance 1 will be used to optimize class 3, misleading the model. Hence, the uncertainty information of the classification score is especially crucial for the performance of the transductive FSL but still under-explored.

In this paper, we propose a novel uncertainty-based method for few-shot learning. It can model the uncertainty of classification results for each query instance and perform uncertainty-aware transductive few-shot learning. 
Specifically, we first augment each query instance with multiple data augmentation methods and classify these augmented instances. 
Then, we compute the mutual information (MI) to measure the uncertainty of the classification scores and calculate the average classification scores for each query instance. 
Next, a weight generator with learnable parameters is used to generate the weights of the average classification scores based on the MI. 
These weights and the average classification scores are then used to calculate the weights of the query instances. The weights of the query instances are used to update the prototype by the weighted average of the query instances and the corresponding support instances.
The refined prototype is affected by the pseudo-classification scores of all query instances and their uncertainty. 

Our contributions can be summarized as follows: 
\begin{enumerate}
    \item To the best of our knowledge, we are the first to measure the uncertainty of the classification result of each query instance by calculating its mutual information from various augmentation algorithms in the transduction few-shot image classification.
    \item We propose a novel method called \ucnA{} (\ucn{}), which refines prototypes by iteratively performing transduction inference based on the pseudo classification results and corresponding mutual information.
    \item The proposed method has been validated on four datasets, and the classification accuracy outperforms other methods by 1\%-18\%. It also achieves state-of-the-art performance in the semi-supervised setting.
\end{enumerate}
\vspace{-0.2in}
\section{Related Work}
\vspace{-0.1in}
\subsection{Transductive few-shot learning}

Compared with inductive few-shot learning methods, transductive few-shot learning methods aim to explore how to utilize the query set to help improve the performance of image classification.
The first type is GNN-based methods that enhance classification performance by exploiting Graph Neural Networks (GNN) in the transductive few-shot learning task.
Typical works include TPN~\cite{propagatelabel}, EGNN~\cite{egnn} and DPGN~\cite{dpgn}.
Some other methods that use query instances to refine prototypes seem more straightforward.
For example, CAN~\cite{crossattention2019} adopts an iterative process during transductive inference. Specifically, in each iteration, the top-k confident query instances are chosen and used to update the prototypes. The uncertainty confidence for each query instance equals the cosine similarity between itself and the prototype.
MCT~\cite{metaconfidence} also adopts the iterative process, but it attempts to adapt the confidence to a specific task by adding an extra learnable temperature to the Euclidean distance metric.

\vspace{-0.1in}
\subsection{Uncertainty-based learning method}
Facts show that uncertainty-based methods have made good progress in the practice of various downstream tasks. 
For example, Yarin et al.~\cite{galuncertainty} were the first to explore the nature of uncertainty in regression and classification tasks extensively, expounded that uncertainty is indispensable, and developed the necessary tools. 
In semi-supervised learning, UPS~\cite{UPS} proposes an uncertainty-aware pseudo-label selection framework to increase the accuracy of pseudo-labeling by reducing the noise during training.
UAFS~\cite{UAFS} proposes a framework based on few-shot image classification, which converts the observed similarity of query-support pairs into probabilistic representations and uses graph convolutional networks (GCN) to take advantage of this uncertainty to achieve more effective optimization results.
TIM~\cite{tim} introduces the concept of mutual information to measure the uncertainty of model prediction results and optimize the model based on the uncertainty of similarity.

\vspace{-0.2in}
\section{Approach}
\label{sec:method}
For different input instances, the output of the neural network has different uncertainties, while the observation noise also has an impact on the uncertainty~\cite{hallucinating}.
Thus, we propose an \ucnA{} (\ucn{}), which models the uncertainty of model classification results and optimizes the prototype by considering the uncertainty of model classification results and classification results.



\vspace{-0.15in}
\subsection{Problem Definition}
\label{sec:prob_def}
To present \ucn{} well, we first introduce some notations about few-shot learning in this section. The training set, validation set and test set are denoted as $\mathcal{D}_{\text {train }}$, $\mathcal{D}_{\text {val }}$ and $\mathcal{D}_{\text {test }}$ respectively. The Few-shot learning method aims to learn a model from the support set $\mathcal{S}$ and use the model to identify sample categories in the query set $\mathcal{Q}$.
Most few-shot learning methods adopt a special paradigm named $N$-way $K$-shot classification to evaluate their performance. To build a $N$-way $K$-shot classification task, we first need to randomly sample $N$ classes from the training set. After that, we not only need to sample $K$ labeled instances to make up the support set $\mathcal{S}$ (\emph{i.e.,} $\mathcal{S} = \left\{ (\mathbf{x}_{i,j}^S, y_{i,j}^S)\mid i=1:N,j=1:K\right\}$) for each chosen class, but also need to sample some instances without labels from these N classes to form the query set $\mathcal{Q}$ (\emph{i.e.,} $\mathcal{Q} = \left\{ \mathbf{x}_i^Q\mid i=1:q  \right\}$, where $q$ is the size of the query set).
In addition, there is no crossover between the samples of the support set $\mathcal{S}$ and query set $\mathcal{Q}$, but they share the label space.
\vspace{-0.1in}
\subsection{Overall Pipeline}
\label{sec:framework}
\begin{figure*}
\begin{center}
\includegraphics[width=0.75\textwidth]{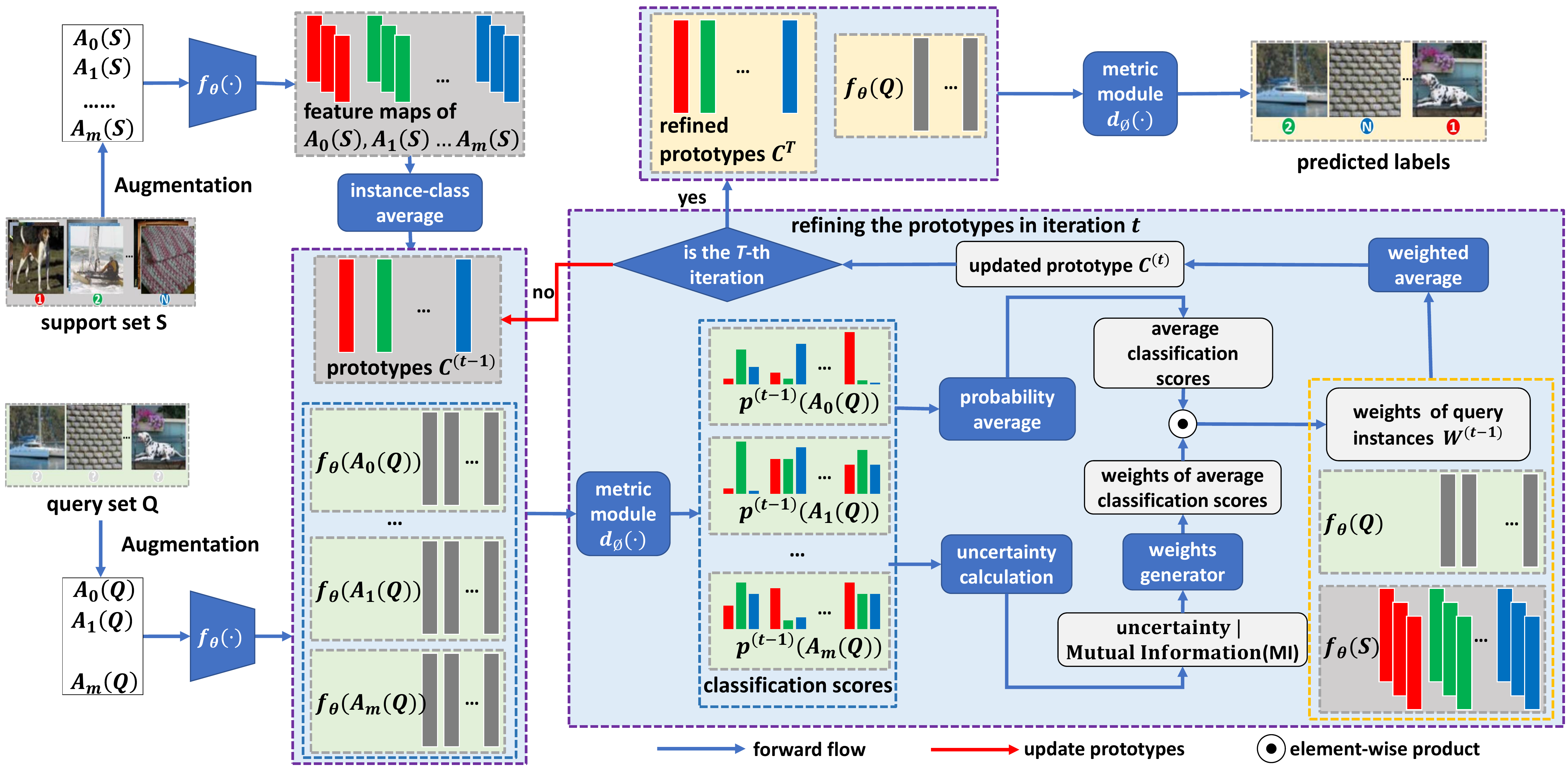}
\end{center}
\vspace{-0.1in}
   \caption{
   \textbf{The framework of the proposed \ucn{}.}
   We optimize the prototype using uncertain information iteratively. First, we model the mutual information (MI) and average classification score for each query instance based on the classification score obtained by applying various augmentations $A(\cdot)$ to support set $\mathcal{S}$ and query set $\mathcal{Q}$. Then there is a process of $T$ iterations to optimize the prototype. For the $t$-th iteration, the MI is fed into a weight generator to calculate the weights of average classification scores, combined with the average classification scores to update the weights of the query instances. The refined prototype for a specific class can be obtained by weighted averaging all corresponding support and query instances.
}
   

\vspace{-0.1in}
\label{fig:framework}
\end{figure*}

This section explains the overview of \ucn{}. The framework of \ucn{} is in Fig.~\ref{fig:framework}.
First, $m+1$ different data augmentation algorithms are applied to each support and query instance as perturbations, leading to $m+1$ different variants for each instance.
For convenience, we use $A_k(\cdot)$ ($0\leq k \leq m$) to denote the data augmentation algorithms, where $A_0(\cdot)$ presents the identity function.
After that, each augmented instance is fed into a feature extractor $f_\theta(\cdot)$ to obtain their feature maps. 
Next, the initial prototypes $\mathcal{C}^{(0)} = \{\mathbf{c}_i^{(0)}\}_{i=1}^N$ are computed by averaging the feature map of these augmented support instances, $i$ indicates the $i$-th class in support set $\mathcal{S}$.
After obtaining the initial prototypes, we perform $T$ consecutive iterations to refine the prototypes.
The $t$-th iteration updates the prototypes based on the feature maps of the augmented instances and the prototypes generated in the previous iteration.
The details of the $t$-th iteration are shown in the section~\ref{ssec:method_refine_proto}. 

\vspace{-0.1in}
\subsection{Refining the Prototypes in Iteration $t$}
\label{ssec:method_refine_proto}
As the $T$ iterations are the core part of \ucn{}, we introduce the details of the $t$-th iteration in this section. 

\textbf{Classify the Augmented Instances.}
To capture the uncertainty of the model prediction results, we employ different data augmentation methods to perturb the prediction results of the model and then calculate the uncertainty of the classification results.

At the beginning of the $t$-th iteration, $m$ data augmentation methods are applied on each instance in the query set $\mathcal{Q}$.
All the augmented query instances are then classified using a metric-based approach, which is shown as: 
\vspace{-1mm}
\begin{footnotesize}
\begin{equation}
    \label{eq:probability}
   {P}^{(t)}(y=i|\mathbf{x}_j^Q, A_k)=\frac{\exp (- d_\phi(f_\theta(A_k(\mathbf{x}_j^Q)), \mathbf{c}_i^{(t-1)}))}
{\sum_{i=1}^N \exp (-d_\phi(f_\theta(A_k(\mathbf{x}_j^Q)), \mathbf{c}_i^{(t-1)}))} ,
\vspace{-1mm}
\end{equation}
\end{footnotesize}
where $\mathbf{x}_j^Q$ denotes $j$-th instance of the query set $\mathcal{Q}$, $\mathbf{c}_i^{(t-1)}$ is the $i$-th prototype in the $(t-1)$-th iteration, and $d_\phi(\cdot,\cdot)$ 
indicates the Euclidean distance metric with a learnable parameter $\phi$, which is formulated as:
\vspace{-1mm}
\begin{footnotesize}
\begin{equation}
    \label{eq:distance_metric}
    d_\phi(\mathbf{x}_1,\mathbf{x}_2) = 
    \left\vert\left\vert\frac{\mathbf{x}_1}{||\mathbf{x}_1||_2 ~ g_\phi(\mathbf{x}_1)}-
    \frac{\mathbf{x}_2}{||\mathbf{x}_2||_2 ~ g_\phi(\mathbf{x}_2)}\right\vert\right\vert_2,
\vspace{-1mm}
\end{equation}
\end{footnotesize}
where $\mathbf{x}_1,\mathbf{x}_2$ represent two different sample, $||.||_2$ is the Euclidean norm, and $g_\phi(\cdot)$
is a temperature generator with learnable parameters. In this paper, we use instance-specific temperatures because, as demonstrated in~\cite{tadam,distillknowledge}, the scaling temperature in a softmax operation can significantly affect the performance of a few learning models. Therefore, an instance-specific temperature adapts the metric to a specific task.

\textbf{Calculate the Average and MI of the Classification Scores.}
The average classification scores $\bar{P}^{(t)}$ of $m+1$ augmented instances for each query sample is first calculated, as:
\vspace{-1mm}
\begin{footnotesize}
\begin{equation}
    \bar{P}^{(t)}(y=i|\mathbf{x}_j^Q) = \frac{1}{m+1}\sum_{k=0}^m {P}^{(t)}(y=i|\mathbf{x}_j^Q, A_k),
\vspace{-1mm}
\end{equation}
\end{footnotesize}
where $i$ is the possible classes of $\mathbf{x}_j^Q$, $j$ indicates the $j$-th element in the query set. 

We draw on the tractable approach proposed by \cite{2011uncertain} to compute mutual information, which approximates the inferred mutual information by adjusting the parameters of the model to obtain multiple predictions.
In this paper, we propose to calculate the uncertainty of the model prediction results with the help of several data augmentation methods. Specifically, MI $\mathbb{I}^{(t)}[y \mid \mathbf{x}_j^Q]$ is calculated based on the classification scores of $m+1$ augmented instances via Shannon entropy $\mathbb{H}(\cdot)$, which is shown as:
\vspace{-1mm}
\begin{footnotesize}
\begin{equation}
    \label{eq:method_mi_2}
    \begin{aligned}
    \mathbb{I}^{(t)} = \mathbb{H}(\bar{P}^{(t)}) - \frac{1}{m+1} \sum_{k=0}^m \mathbb{H}({P}^{(t)}).
    \end{aligned}
\vspace{-1mm}
\end{equation}
\end{footnotesize}

\textbf{Infer Weights for Query Instances.} 
This paper proposes an uncertainty-aware few-shot learning method, which considers uncertainty information and instance classification scores to select appropriate query instances. 
All the uncertainty information of query instances is first fed to a weight generator $h_\psi(\cdot)$, which is a multi-head attention module in this paper.
Then, the outputs are applied on the average classification scores to get the weight matrix $\mathbf{W}^{(t)}$ of query instances, as Eq.~\ref{eq:method_weight_generator_2}, to update the prototypes. The weight matrix $\mathbf{W}^{(t)}$ integrates mutual information and probability distributions through multi-head attention mechanism and provides a hint on the uncertainty of different query instances since the learned weights $\mathbf{W}^{(t)}$ reflect the impact of uncertainty on the query instances.
\vspace{-1mm}
\begin{footnotesize}
\begin{equation}
    \begin{aligned}
    \label{eq:method_weight_generator_2}
        \mathbf{W}^{(t)} &=\left[h_\psi\left(\mathbb{I}^{(t)}\right)\odot \bar{P}^{(t)}\right] = \left[w_{i,j}^{(t)}\right],
    \end{aligned}
\vspace{-1mm}
\end{equation}
\end{footnotesize}
where $\odot$ represents element-wise product. The weight generator $h_\psi(\cdot)$ is constructed by one multi-head attention module, as the multi-head attention can discover the relationship among elements of the input sequence.

\textbf{Optimize Prototypes.}
$\mathbf{W}^{(t)}$ is used to update the prototypes, which are linear combinations of the embedding of the support instance and the query instance with a fixed weight of $1$ for the support instances and a weight $\mathbf{W}^{(t)}$ for the query instances.
The updated prototypes are in Eq.~\ref{eq:method_weight_generator_3}.
\vspace{-1mm}
\begin{footnotesize}
\begin{equation}
    \label{eq:method_weight_generator_3}
    \mathbf{c}_i^{(t)} = \frac{\sum_{j=1}^K 1 \cdot \mathbf{x}_{i,j}^S + \sum_{j=1}^q w^{(t)}_{i,j} \cdot \mathbf{x}_j^Q}{K+\sum_{j=1}^q w^{(t)}_{i,j}}.
\vspace{-1mm}
\end{equation}
\end{footnotesize}
The prototype $\mathbf{c}_i^{t}$ is updated with a weighted average using the above weights $w^{(t)}_{i,j}$ combined with query instances and support instances. When selecting a query sample for prototype optimization, we consider the uncertainty of the classification result and the classification result for this query sample.
\vspace{-0.1in}
\subsection{Loss Function}
\label{ssec:method_objective}
In this work, a joint loss function $L$ is proposed to train \ucn{}. It combines a classification loss $L_{\text{cls}}$ and a regularization loss $L_{\text{gen}}$ as Eq.~\ref{eq:loss_main}.
\vspace{-1mm}
\begin{footnotesize}
\begin{equation}
    \label{eq:loss_main}
    L = L_{\text{cls}} + \lambda L_{\text{gen}},
\vspace{-1mm}
\end{equation}
\end{footnotesize}
where $\lambda$ is the hyperparameter to balance two losses.
We empirically set it to 0.5 in the experiment.

The classification loss $L_{\text{cls}}$ is the cross-entropy loss of the classification scores of all non-augmented query instances, and the goal of $L_{\text{cls}}$ is to increase the classification accuracy of the model.
The regularization loss $L_{\text{gen}}$ aims to regularize the weight generator. The outputs of the support instances are as close as possible to their one-hot labels since the labels of support instances are already known. The regularization loss $L_{\text{gen}}$ is formulated as Eq.~\ref{eq:loss_reg_4}.
\vspace{-1mm}
\begin{footnotesize}
\begin{equation}
    \label{eq:loss_reg_4}
    L_{\text{gen}} = -\frac{1}{NK}  \sum_{i=1}^N \sum_{j=1}^K \text{BCE}\left(w^{(T)}_{i,j}, \text{onehot}(i, N)\right),
\vspace{-1mm}
\end{equation}
\end{footnotesize}
where $\text{BCE}(\cdot,\cdot)$ indicates the binary cross entropy loss, $w^{(t)}_{i,j}$ denotes the weight of the $j$-th supported instance in the $i$-th class inferred from the weight generator $h_\psi(\cdot)$.
$N$ is the class number in support set, $K$ is the instance number in each class, and $\text{onehot}(i, N)$ indicates the one-hot vector whose length equals to $N$ with its $i$-th position (starts from $1$) being $1$. 
\vspace{-0.2in}
\section{Experiments}
\label{sec:experiments}

\subsection{Datasets, Backbone and Implementation Details}

\label{ssec:experiments_datasets}
To validate \ucn{}, we conduct extensive experiments on four public datasets: \miniimagenet{}~\cite{matching}, \tieredimagenet{}~\cite{tieredimagenet}, FC100~\cite{tadam}, and CIFAR-FS{}~\cite{cifarfs}. We choose different backbones (\emph{e.g.,} Conv-64/256, Resnet12/18 and WRN-28-10) for evaluation. More implementation details and experimental results are shown in the supplemental material.


\begin{table}[t]
    \tiny
    \setlength{\tabcolsep}{2mm}
	\centering
	\fontsize{8}{11}\selectfont   
	\caption{Classification accuracy on \miniimagenet{} and \tieredimagenet{} reported with 95\% confidence intervals. 
	"FE" column describes the backbone, "Tr" indicates whether methods are transductive, and the partially transductive methods are marked as "BN" because of using the mean and variance of query instance batches in BatchNorm layers~\cite{egnn,dpgn}.
	The superscript $\dagger$ indicates dense classification~\cite{denseclassification} is adopted over feature maps.
	C64/256 is for Conv-64/256, R12/18 for ResNet-12/18 and W28 for WRN-28-10.}
	\renewcommand{\arraystretch}{0.5}
	\begin{tabular}{p{1.4cm}p{0.4cm}p{0.3cm}p{0.92cm}p{0.92cm}p{0.92cm}p{0.92cm}p{0.92cm} }
    \toprule
    		\multirow{2}{*}{Method} &   \multirow{2}{*}{FE}    &\multirow{2}{*}{Tr}               &\multicolumn{2}{c}{\miniimagenet{}(\%)} 
    		&\multicolumn{2}{c}{\tieredimagenet{}(\%)}\\
		\cmidrule(lr){4-7}  &   &   &   1-shot &    5-shot 
		                             &   1-shot &    5-shot\\
	
	\midrule
    ProtoNet~\cite{protonet}    &   C64 &   No  &   49.42\tiny{$\pm$0.78}  &   68.20\tiny{$\pm$0.66}  &   53.31\tiny{$\pm$0.89}  &   72.69\tiny{$\pm$0.74}\\
    MatchNet~\cite{matching}    &   C64 &   No  &   43.56\tiny{$\pm$0.84}  &   55.31\tiny{$\pm$0.73}  &   -  &   -\\
    MAML~\cite{maml}            &   C64 &   BN  &   48.70\tiny{$\pm$1.84}  &   63.15\tiny{$\pm$0.91}  &   -  &   - \\
    RN~\cite{rn}                &   C64 &   BN  &   50.44\tiny{$\pm$0.82}  &   65.32\tiny{$\pm$0.70}  &   54.48\tiny{$\pm$0.93}  &   71.32\tiny{$\pm$0.78} \\
    TPN~\cite{propagatelabel}   &   C64 &   Yes &   53.75 &   69.43       &   57.53 &   72.85\\
    Ent~\cite{ent}              &   C64 &   Yes &   50.46\tiny{$\pm$0.62}  & 66.68\tiny{$\pm$0.52}     &   58.05\tiny{$\pm$0.68}  & 74.24\tiny{$\pm$0.56} \\
    \rowcolor{gray!40}\ucn{}      &   C64 &   Yes &   \textbf{57.97\tiny{$\pm$0.66}}  & \textbf{73.01\tiny{$\pm$0.75}} 
                                                &   \textbf{58.95\tiny{$\pm$1.13}}  & \textbf{75.82\tiny{$\pm$0.79}} \\
    \midrule
    EGNN~\cite{egnn}            &   C256 &   Yes &  -  & 76.37 &   -  & 70.15  \\
    \rowcolor{gray!40}\ucn{}       &   C256 &   Yes &   \textbf{59.19\tiny{$\pm$1.14}}  & \textbf{77.70\tiny{$\pm$0.64}} 
                                                &   \textbf{68.90\tiny{$\pm$1.17}}  & \textbf{82.78\tiny{$\pm$0.67}} \\
    
    \midrule
    TADAM~\cite{tadam}          &   R12 &   No  &   58.5~~\tiny{$\pm$~0.3}    &   76.7~~\tiny{$\pm$~0.3} &   -  &  - \\
    CAN~\cite{crossattention2019} & R12 & No    &   63.85\tiny{$\pm$0.48}      &   79.44\tiny{$\pm$0.34}   
                                                &   69.89\tiny{$\pm$0.51}  & 84.23\tiny{$\pm$0.37} \\
    DeepEMD~\cite{deepemd}        & R12 & No    &   65.91\tiny{$\pm$0.82}      &   82.41\tiny{$\pm$0.56}   
                                                &   71.16\tiny{$\pm$0.87}  & 86.03\tiny{$\pm$0.58}  \\
    FEAT~\cite{feat}            &   R12 & No    &   66.78\tiny{$\pm$0.20}      &   82.05\tiny{$\pm$0.14}   
                                                &   70.80\tiny{$\pm$0.23}  & 84.79\tiny{$\pm$0.16}  \\

    CAN~\cite{crossattention2019} 
                                &   R12 & Yes   &   67.19\tiny{$\pm$0.55}      &   80.64\tiny{$\pm$0.35} 
                                                &   73.21\tiny{$\pm$0.58}  & 84.93\tiny{$\pm$0.38} \\
    DPGN~\cite{dpgn}            &   R12 & Yes   &   67.77\tiny{$\pm$0.32}      &   84.60\tiny{$\pm$0.43} 
                                                &   72.45\tiny{$\pm$0.51}  & 87.24\tiny{$\pm$0.39}   \\
    EPNet~\cite{epnet}          &   R12 & Yes   &   66.50\tiny{$\pm$0.89}      &   81.06\tiny{$\pm$0.60} 
                                                &   76.53\tiny{$\pm$0.87}  & 87.32\tiny{$\pm$0.64} \\
    Ent~\cite{ent}           &   R12 & Yes   &   62.35\tiny{$\pm$0.66}  &   74.53\tiny{$\pm$0.54}
                                                &   68.41\tiny{$\pm$0.73}  &   83.41\tiny{$\pm$0.52}\\
    IEPT\cite{IEPT}           &   R12 & Yes   &   67.05\tiny{$\pm$0.44}  &   82.90\tiny{$\pm$0.30}
                                                &   72.24\tiny{$\pm$0.50}  &   86.73\tiny{$\pm$0.34} \\
    UAFS\cite{UAFS}           &   R12 & Yes   &   64.22\tiny{$\pm$0.67}  &   79.99\tiny{$\pm$0.49}
                                                &   69.13\tiny{$\pm$0.84}  &   84.33\tiny{$\pm$0.59} \\
    \rowcolor{gray!40}\ucn{}       &   R12 & Yes   
    & \textbf{76.78\tiny{$\pm$1.01}}& \textbf{85.35\tiny{$\pm$0.61}} 
    & \textbf{80.00\tiny{$\pm$1.12}} & \textbf{87.50\tiny{$\pm$0.64}}\\
    MCT$^{\dagger}$~\cite{metaconfidence} & R12 & Yes
                                        & 78.55\tiny{$\pm$0.86}        & 86.03\tiny{$\pm$0.42} 
                                        & \textbf{82.32\tiny{$\pm$0.81}}        & 87.36\tiny{$\pm$0.50}\\   
    \rowcolor{gray!40} \ucn{}$^{\dagger}$ & R12 & Yes 
                                        & \textbf{86.40\tiny{$\pm$0.46}}        & \textbf{87.22\tiny{$\pm$0.43}} 
                                        & 82.02\tiny{$\pm$1.01}        & \textbf{90.83\tiny{$\pm$0.39}} \\               
    \midrule
    
    CTM~\cite{ctm}              &   R18 & No    &   64.12\tiny{$\pm$0.82}     &   80.51\tiny{$\pm$0.13}  &   68.41\tiny{$\pm$0.39}  & 84.24\tiny{$\pm$1.73} \\
    \rowcolor{gray!40}\ucn{}      &   R18 & Yes   & \textbf{73.98\tiny{$\pm$0.94}}& \textbf{83.94\tiny{$\pm$0.59}} 
                                                & \textbf{75.79\tiny{$\pm$1.03}} & \textbf{87.02\tiny{$\pm$0.43}} \\
    \midrule
    CC+rot~\cite{ccrot}         &   W28 &  Yes  & 62.93\tiny{$\pm$0.45}  & 79.87\tiny{$\pm$0.33}        
                                                & 70.53\tiny{$\pm$0.51}  & 84.98\tiny{$\pm$0.36} \\
    
    EPNet~\cite{epnet}              &   W28 &  Yes  & 70.74\tiny{$\pm$0.85}  & 84.34\tiny{$\pm$0.53}        
                                                &   78.50\tiny{$\pm$0.91}  & 88.36\tiny{$\pm$0.57}  \\
    SIB~\cite{sib}              &   W28 &  Yes  &   70.0~~\tiny{$\pm$0.6~~}  & 79.2~~\tiny{$\pm$0.4~~}        &   -  & -  \\       
    E$^{3}$BM~\cite{ebm}        &   W28 & Yes   &   71.4~~\tiny{$\pm$0.5~~}  &   81.2~~\tiny{$\pm$0.4~~}
                                                &   75.6~~\tiny{$\pm$0.6~~}  &   84.3~~\tiny{$\pm$0.4~~}\\
    \rowcolor{gray!40}\ucn{}  &   W28 & Yes   & \textbf{78.55\tiny{$\pm$0.97}}  & \textbf{85.99\tiny{$\pm$0.57} }
                                          & \textbf{79.66\tiny{$\pm$1.02}}  & \textbf{88.87\tiny{$\pm$0.63}} \\
    \bottomrule
	\end{tabular}
	\label{table:sota_mini_tire}
    \vspace{-0.25in}
\end{table}   
\vspace{-0.2in}
\subsection{Comparison with State-of-the-art Methods}
\label{ssec:experiments_sota}


\begin{table}[t]
\vspace{-0.25in}
	\centering
	\fontsize{8}{11}\selectfont   
    \caption{Classification accuracy on FC100 and CIFAR-FS are reported with 95\% confidence intervals. }
    \renewcommand{\arraystretch}{0.5}
	\begin{tabular}{p{1.4cm}p{0.4cm}p{0.3cm} p{0.92cm} p{0.92cm} p{0.92cm} p{0.92cm} p{0.92cm} }
		\toprule
		\multirow{2}{*}{Method} &\multirow{2}{*}{FE}    &\multirow{2}{*}{Tr}   &\multicolumn{2}{c}{FC100(\%)} &\multicolumn{2}{c}{CIFAR-FS(\%)} \\
		\cmidrule(lr){4-7}  &   &   &   1-shot &    5-shot &   1-shot &    5-shot \\
	\midrule
    TADAM~\cite{tadam}                          &   R12 &   No  &   40.1~~ \tiny{$\pm$0.4~~}    &   56.1~~ \tiny{$\pm$0.4~~} 
                                                                &   -  &   - \\
    DeepEMD~\cite{deepemd}                      &   R12 &   No  &   46.47\tiny{$\pm$0.48}      &   \textbf{63.22\tiny{$\pm$0.71}} 
                                                                &   -  &   - \\
    DPGN~\cite{dpgn}                            &   R12     &   Yes &   -  &   -
                                                                &   77.9~~\tiny{$\pm$0.5}&   90.2~~~\tiny{$\pm$0.4} \\
    UAFS\cite{UAFS}           &   R12 & Yes   &      41.99\tiny{$\pm$0.58}  &   57.43\tiny{$\pm$0.38} &   74.08\tiny{$\pm$0.72}  &   85.92\tiny{$\pm$0.42}\\
    \rowcolor{gray!40}\ucn{}                       &   R12 &   Yes & \textbf{48.00\tiny{$\pm$1.04}}&   60.39\tiny{$\pm$0.82}
                                                                 &   \textbf{88.90\tiny{$\pm$0.95}}  & \textbf{92.18\tiny{$\pm$0.63}} \\
    
    \midrule
    SIB~\cite{sib}                              &   W28 &   Yes &   -  & -    &   80.0~~~\tiny{$\pm$0.6}    &   85.3~~~\tiny{$\pm$0.4} \\
    \rowcolor{gray!40}\ucn{}           &   W28 &   Yes & \textbf{47.61\tiny{$\pm$0.99}}&   59.74\tiny{$\pm$0.78} 
                                                                 &   \textbf{85.64\tiny{$\pm$0.93}}  & \textbf{88.90\tiny{$\pm$0.58}} \\
    \bottomrule
    \end{tabular}
\label{table:sota_fc100_cifar}
\end{table}

\begin{figure}[t]
\centering  
\subfigure[MCT]{
\label{subfig:MCT_boundary}
\includegraphics[width=0.2\textwidth]{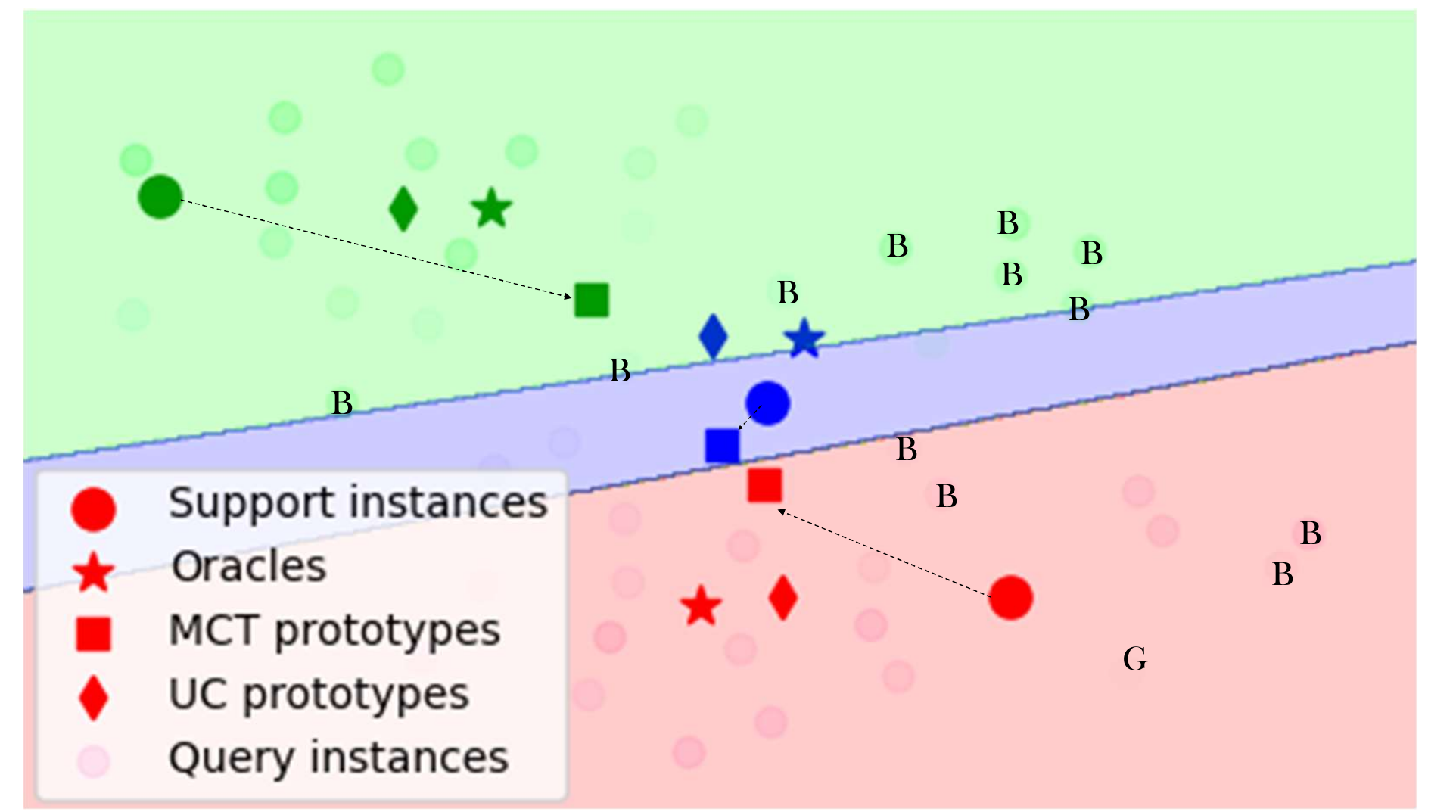}}
\subfigure[\ucn{}]{
\label{subfig:UCN_boundary}
\includegraphics[width=0.2\textwidth]{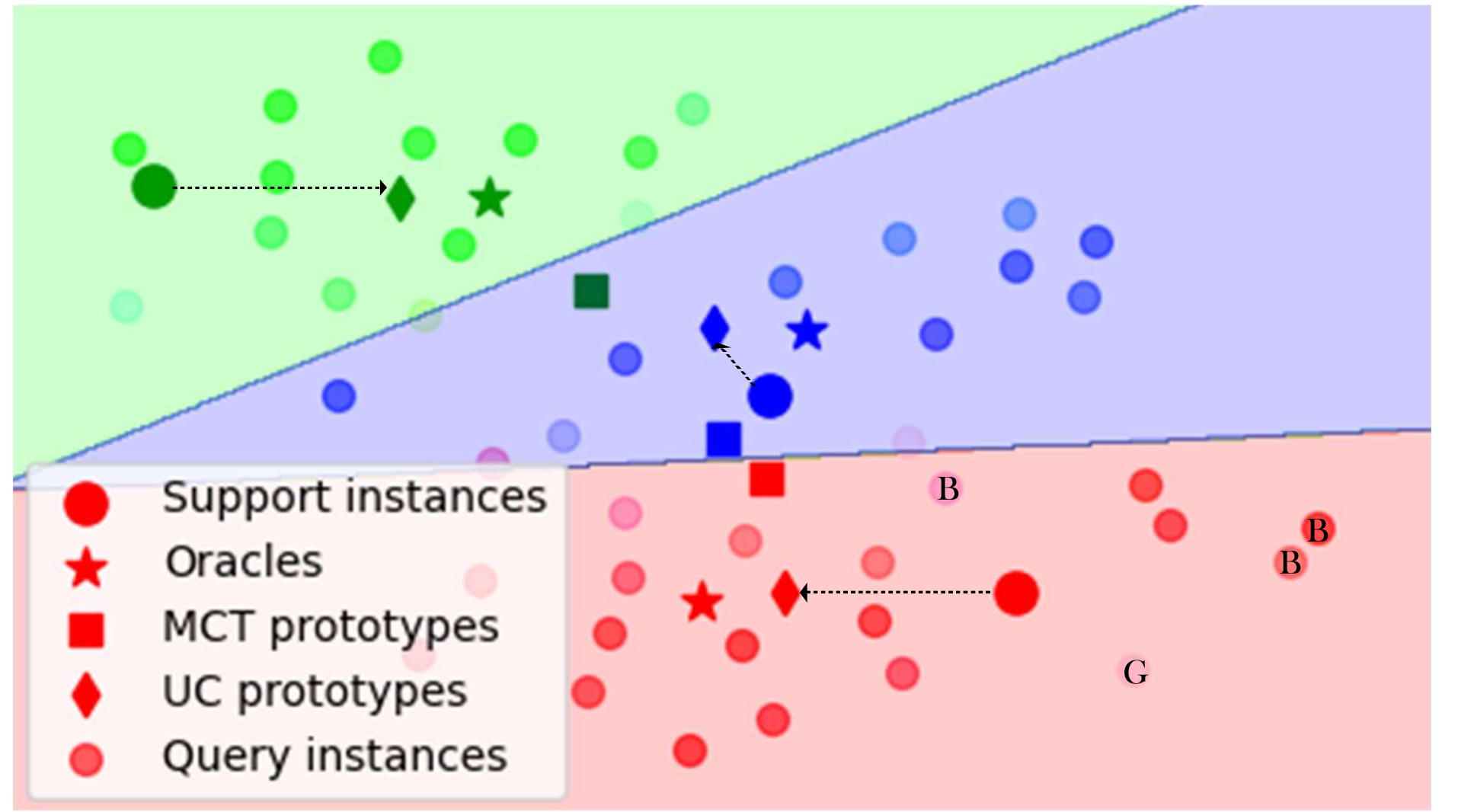}}
\caption{t-SNE figures for the results produced by MCT~\cite{metaconfidence} and our \ucn{} on 3-way 1-shot tasks. 
Classes are represented by red, green, and blue, respectively.
The closer to Oracles, the better the prototype.
The transparency of query instances means the maximum weight it gets. 
}
\vspace{-0.23in}
\label{fig:main_tsne}
\end{figure}

\textbf{Quantitative Comparison.} 
Table~\ref{table:sota_mini_tire},~\ref{table:sota_fc100_cifar} show that \ucn{} surpasses most of the methods on all datasets and demonstrates the effectiveness of \ucn{}. 
It shows that the transductive FSL learning methods work better than the inductive FSL learning methods, indicating that query samples benefit prototype optimization. Note that the R12 backbone and pre-train strategy of the proposed UNC are most similar to FEAT~\cite{feat}. Compare to FEAT, \ucn{} achieves significant improvement on \miniimagenet{} and \tieredimagenet{} in 1-shot accuracy. 
Unlike UAFS~\cite{UAFS} which model uncertainty of the similarities of query support pairs, we propose a new way of computing uncertainty information of query instances by computing the mutual information of the augmented query instances. 
Moreover, the obtained uncertainty information and the classification results of the query instances are used to guide the transductive inference process.
Relative to UAFS, our prediction accuracy increased by a large margin on all metrics. In addition, we obtain competitive results in a semi-supervised setting and present the results in supplementary materials.


\begin{table}[t]
\vspace{-0.25in}
	\centering
    \fontsize{8}{11}\selectfont   
    \caption{Ablation study of uncertainty information (UI) and regularization loss ($L_{gen}$) on \miniimagenet{}, measured in \%.}
    \renewcommand{\arraystretch}{0.5}
    
    \begin{tabular}{llllll}
		\toprule
        Method & UI & $L_{gen}$ & Tr & 1-shot & 5-shot\\		
		\midrule
baseline &  \XSolidBrush   &  \XSolidBrush   &\XSolidBrush    & 65.13\tiny{$\pm$0.80}         & 82.29\tiny{$\pm$0.53} \\ 
\ucn{}\_A & \XSolidBrush  & \XSolidBrush   & \checkmark    & 69.65\tiny{$\pm$1.10}            & 84.21\tiny{$\pm$0.55} \\ 
\ucn{}\_B & \checkmark  &  \XSolidBrush & \checkmark
                                & 73.92\tiny{$\pm$1.06}            & 84.20\tiny{$\pm$0.57} \\
\rowcolor{gray!40}\ucn{} &  \checkmark &\checkmark  & \checkmark
                                & \textbf{76.78\tiny{$\pm$1.01}}   & \textbf{85.35\tiny{$\pm$0.61}}\\ 
\bottomrule
\end{tabular}
\label{table:ablation_uc}
\end{table}

\textbf{Qualitative Comparison.} To analyze the effectiveness of \ucn{}, we compare it with the previous state-of-the-art method MCT~\cite{metaconfidence} based on the prototype refining strategy.
Figure~\ref{fig:main_tsne} visualized ProtoNet-style~\cite{protonet} prototype (support instances), instance feature maps (query instances), refined prototype generated by our method (UC prototype), MCT prototype, and the oracle prototype. 
The arrows mark the changes made to the prototype by the MCT and \ucn{} methods.
The oracle prototype is obtained by averaging the features of all samples of the same class in the dataset, 
which can represent the commonality of instances in a category because the averaging operation neutralizes the random noise over a wide range of instances.
The RGB value of the points for query instances $\mathbf{x}_j^Q$ is determined by the proportion among the weights $w_{0,j}^{(T)}$, $w_{1,j}^{(T)}$ and $w_{2,j}^{(T)}$, and its transparency reflects the maximum weight of the corresponding query instance, that is $\max (w_{0,j}^{(T)}, w_{1,j}^{(T)}, w_{2,j}^{(T)})$.
In this way, we can intuitively understand the contribution of each query instance to the refined prototypes.
In addition, Figure~\ref{subfig:MCT_boundary},~\ref{subfig:UCN_boundary} also draw the approximate decision boundaries based UC prototypes and MCT prototypes, respectively. 
All misclassified query instances are also marked with ground truth labels.

From Fig.~\ref{fig:main_tsne}, we can see that compared with MCT, the prototype generated by \ucn{} is closer to the ideal state (\emph{i.e.}, the oracle prototype).
The query instances of the same category have higher weights, while the misclassified instances often have lower weights. This result reveals that \ucn{} can make the decision boundary more reasonable.
These phenomena are consistent with our hypothesis that using the uncertainty information of classification score can improve the prototype in transductive few-shot learning.

\vspace{-0.1in}
\subsection{Ablation Study}
\label{ssec:experiments_abl}
We perform ablation experiments on several components of \ucn{} to study their effect, including the impact of uncertainty information, the effect of $T$, the choice of augmentations, and the structure of the weight generator $h_\psi(\cdot)$. Due to space limitations, some of these ablation experiments are in the supplemental material. 

\begin{figure}[t] 
    \centering
    \includegraphics[width=0.3\textwidth]{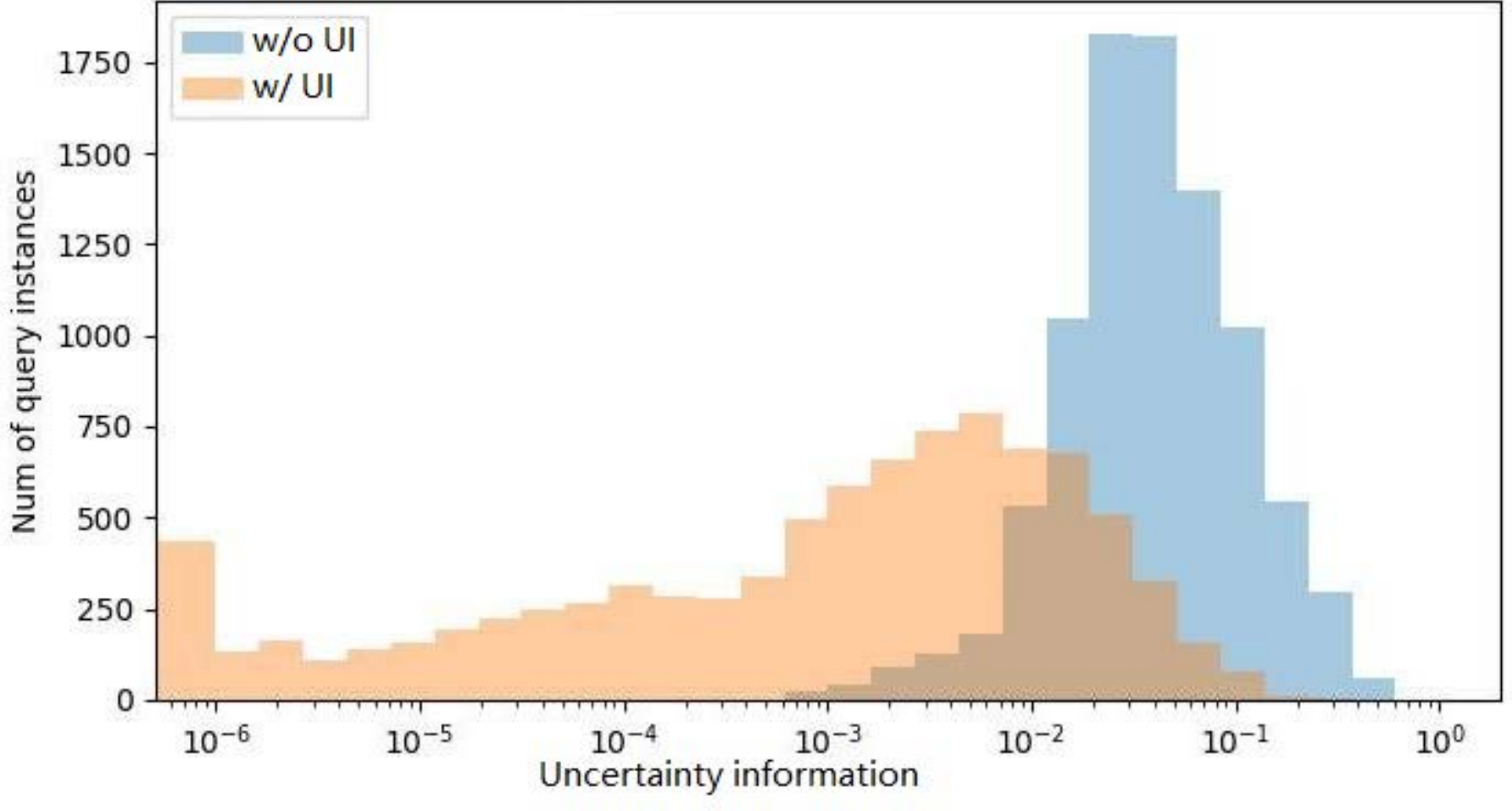}
    \caption{Distributions of uncertainty information of classification scores.}
     \vspace{-0.23in}
    \label{fig:abl_distribution} 
\end{figure}

In Table \ref{table:ablation_uc}, we conduct an ablation study of the uncertainty information (UI) and the proposed regularization loss function ($L_{gen}$) in \miniimagenet{}. 
The baseline represents the model of ProtoNet-style~\cite{protonet}, and it also takes the average of the corresponding augmented instances as the basis. \ucn{}\_A and \ucn{}\_B have the same structure and parameters as \ucn{}, except for some differences in UI and loss functions, where \ucn{}\_A directly uses the classification score of query instances as UI. 
Table \ref{table:ablation_uc} illustrates the effectiveness of the proposed UI. 
It shows that with the help of UI, the accuracy of the experiment increases by 4.27\% for the 5-way 1-shot setting.
We argue that the uncertainty of the model prediction results is high in the 5-way 1-shot setting. More beneficial instances in the query set can be selected to optimize the prototypes after taking the uncertainty information into account. Table \ref{table:ablation_uc} also demonstrates the effectiveness of $L_{gen}$, especially on 5-way 1-shot setting.

In Fig.~\ref{fig:abl_distribution}, we conduct 3,000 5-way 1-shot episodes using the UCN\_A (made with w/o UI) and UCN\_B (marked with w/UI) and collect the top three query instances in each episode (ordered by their weights in class 0).
The distribution shows that samples with a lower uncertainty in classification scores usually receive relatively higher weights when uncertainty information is introduced into the prototype refinement. 

\vspace{-0.2in}
\section{Conclusion}
\label{sec:conclusion}
In this paper, we propose an Uncertainty-based network for few-shot learning, which utilizes the uncertainty information of query instances to refine the prototypes through transductive inference. 
Notably, we compute the weights of all the query instances based on both the classification scores and the uncertainty information of these scores. Extensive experiments on various benchmarks show that \ucn{} outperforms the state-of-the-arts significantly on both transductive few-shot learning and semi-supervised few-shot learning tasks. 

\bibliographystyle{IEEEbib}
\bibliography{icme2022template}

\end{document}


\sloppy

\title{Supplementary Material for Uncertainty-based Network for Few-shot Image Classification}
\name{Minglei Yuan, Qian Xu, Chunhao Cai, Yin-Dong Zheng, Tao Wang, Tong Lu$^{*}$ and Wenbin Li}
\address{National Key Laboratory for Novel Software Technology, Nanjing University \\
         \{mlyuan, chinyxu\}@smail.nju.edu.cn, chunhao.cai@smartmore.com, \\
         \{ydzheng0331, taowangzj\}@gmail.com, \{lutong, liwenbin\}@nju.edu.cn}

\maketitle

\ificcvfinal\thispagestyle{empty}\fi
\section{Summarizing of the Paper}
In this paper, we propose a \ucnA{} (UCN) method, which models the uncertainty of classification results with the help of mutual information.
Specifically, we first data augment and classify the query instance and calculate the mutual information of these classification scores.
Then, mutual information is used as uncertainty to assign weights to classification scores, and the iterative update strategy based on classification scores and uncertainties assigns the optimal weights to query instances in prototype optimization.

\section{Datasets Details}
\label{smsec:datasets}
In order to demonstrate the versatility of \ucn, we conduct experiments on four different datasets : \textbf{\miniimagenet{}}~\cite{matching}, \textbf{\tieredimagenet{}}~\cite{tieredimagenet}, \textbf{FC100}~\cite{tadam}, and \textbf{CIFAR-FS{}}~\cite{cifarfs}.

\textbf{\miniimagenet}~\cite{matching} is derived from ImageNet~\cite{imagenet}, it contains 100 classes and each class has 600 instances. The size of all instances is 84$\times$84. Following the practice of~\cite{rn}, we use the split of 64/16/20 classes for the training/validation/testing set.

\textbf{\tieredimagenet}~\cite{tieredimagenet} is also a subset of ImageNet, which is relatively larger than \miniimagenet{} though. It consists of 779,165 images collected from 34 superclasses, each superclass consisting of several low-level classes similar in spirit. We split the dataset into 20/6/8 superclasses or 351/97/160 low-level classes for the training/validation/testing set. All the images are resized to 84$\times$84 too.

\textbf{FC100}~\cite{tadam} is a variant of the well-known CIFAR-100 dataset~\cite{cifar100}, which is specifically designed for few-shot classification. Images in FC100 have a size of 32$\times$32, as the original CIFAR-100 does. FC100 contains 20 superclasses, similar to \tieredimagenet. Our split is 12/4/4 superclasses or 60/20/20 low-level classes for the training/validation/testing set.

\textbf{CIFAR-FS}~\cite{cifarfs} is a subset of CIFAR-100 as well. It contains 100 classes that describe general object categories, each containing 600 images with the size of 32$\times$32. 
We split the dataset into 64/16/20 classes for training/validation/testing according to the splitting method in~\cite{metaconfidence}.
\begin{table*}[!htbp]
	\centering
	\fontsize{8}{11}\selectfont   
    \caption{Semi-supervised results on \miniimagenet{} (mini) and \tieredimagenet{} (tiered). All results are based on pre-trained ResNet-12 with the full dataset in a traditional supervised manner. $^\ddagger$ denotes the results reported in ~\cite{LST}. w/ $\mathcal{D}$ means that the unlabeled set includes 3 distracting classes which do not overlap with the label space of the support set }
    \renewcommand{\arraystretch}{0.5}
    {
    \resizebox{0.75\textwidth}{!}{
	\begin{tabular}{lcccccccc}
		\toprule
		\multirow{2}{*}{Method} &\multicolumn{2}{c}{mini}  & \multicolumn{2}{c}{tiered} & \multicolumn{2}{c}{mini w/ $\mathcal{D}$} & \multicolumn{2}{c}{tiered w/ $\mathcal{D}$}\\
		& 1-shot & 5-shot & 1-shot & 5-shot & 1-shot & 5-shot & 1-shot & 5-shot \\
		\midrule
        Masked Soft $k$-means w/ MTL$^\ddagger$~\cite{tieredimagenet} 
        & 62.1                  & 73.6    
        & 68.6                  & 81.0 
        & 61.0                  & 72.0  
        & 66.9                  & 82.2    \\
        TPN w/ MTL$^\ddagger$~\cite{propagatelabel} 
        & 62.7                  & 74.2
        & 72.1                  & 83.3
        & 61.3                  & 72.4
        & 71.5                  & 82.7
        \\
        LST~\cite{LST}  
        & 70.1                  & 78.7
        & 77.7                  & 85.2
        & 64.1                  & 77.4
        & 73.5                  & \textbf{83.4}
        \\
        MCT~\cite{metaconfidence} 
        & 73.8                  & 84.4
        & 76.9                  & 86.3
        & 69.6                  & 81.3
        & 74.5                  & 84.0
        \\
        \rowcolor{gray!40}
        \ucn{} w/o DC
        & \textbf{77.3}        & \textbf{85.4}
        & \textbf{80.1}        & \textbf{87.4}
        & \textbf{70.0}        & \textbf{80.6}
        & \textbf{74.7}        & 81.8
        \\

\bottomrule
\end{tabular}}
}
\vspace{-0.2in}
\label{table:sota_ssl_mini}
\end{table*}

\section{Implementation Details}
\label{smsec:network_arch_detail}
\subsection{Backbone}
For fair comparisons, five backbone networks are adopted in the feature extractor $f_\theta(\cdot)$ of \ucn{}. Specifically, these backbone networks are Conv64, Conv256, ResNet-12, ResNet-18 and WRN-28-10.

\textbf{Conv64 and Conv256} are both consisted of four convolutional layers, with a BatchNorm~\cite{batchnorm} layer, a 2$\times$2 MaxPool layer and a ReLU~\cite{relu} activation function after each convolutional layer. They are different only in the number of filters of each convolutional layer: Conv64 has 64/64/64/64 filters, while Conv256 has 64/96/128/256 filters. 

\textbf{ResNet-12 and ResNet-18} are both variants of the popular ResNet~\cite{resnet}. The structure of ResNet-18 is already presented in~\cite{resnet}. ResNet-12 consists of four residual blocks~\cite{resnet}, each containing three 3$\times$3 convolutional layers. Following each convolutional layer are a BatchNorm layer and a LeakyReLU~\cite{leakyrelu} activation function (with a negative slope of 0.1). At the end of each residual block, a 2$\times$2 max-pooling is inserted after the shortcut connection. 

\textbf{WRN-28-10} is a variant of the wide residual network~\cite{wrn}, which is also adopted as a feature extractor in some recent works. It is built with 12 wide residual basic blocks~\cite{wrn} and has a wide factor of 10 and a depth of 28. 

\subsection{Temperature Generator}
The temperature generator $g_\phi(\cdot)$ is implemented by a fully connected network including two hidden layers. The output sizes of the hidden layers are 512 and 128, respectively. The dimensions of the keys and the values in the MultiAtt($\cdot$) are both set to 128.



\subsection{Network Training, Evaluation and Hyperparameters}
\label{sssec:experiments_impl_train}
The meta-training process of \ucn{} lasts 50,000 episodes. We use the SGD optimizer with Nesterov momentum~\cite{nesterov} in all experiments, and the initial learning rate of the feature extractor is 0.0001, while others are 0.001. 

The meta-testing process follows the previous method~\cite{matching}, which samples 600 random episodes and averages the accuracy of these episodes. The 95\% confidence intervals will be reported, as well as the mean accuracy.
We sample 15 query instances for each class in an episode, while the number is reduced to 8 when WRN-28-10 is used as the backbone due to GPU memory limitations. 
\vspace{-2mm}
\begin{equation}
    \label{eq:loss_main}
    L = L_{\text{cls}} + \lambda L_{\text{gen}},
\end{equation}
where $\lambda$ is a hyperparameter given by experience. We empirically set it to 0.5 in the experiment.

The validation phase uses the same protocol, except that the data set is the validation set. In addition, we adopt the number of iterations $T=6$ and $\lambda=0.5$ in Eq.~\ref{eq:loss_main}. Horizontal flipping, histogram equalization, color adjustment (factor=3) and random cropping (crop size=0.68$\times$original size) are used as data enhancement algorithms. A multi-headed attention module is used. The choice of these hyperparameters will be discussed in ablation experiments.



\section{Pretraining}
\label{smsec:pretraining}
When using Conv64, Conv256, ResNet-12, ResNet-18 and WRN-28-10 as the feature extractor, we follow the practice adopted in most recent works~\cite{tadam,ctm,dpgn,deepemd,egnn,sib,epnet,crossattention2019} and pretrain the feature extractor before the meta-learning stage. This is accomplished through a general classification task on the training split of the dataset. 

The whole pretraining process lasts 200 epochs. We validate the model on a randomly sampled 5-way 1-shot task from the validation split at the end of each epoch. We use SGD as the optimizer. The initial learning rate is set to $0.1$, and is multiplied by $\sqrt{\frac{1}{10}}$ every time when the validation accuracy doesn't improve in the last consecutive seven epochs. We perform 6,000 iterations of learning rate warm-up at the beginning of pretraining. The learning rate increases at a linear pace during warm-up.

\section{Additional Results}
\subsection{Semi-supervised Inference}
\label{smssec:ssl_results}
 To further validate the effectiveness and generality of our \ucn{}, 5-way 5-shot and 5-way 1-shot experiments are conducted in a semi-supervised setting. We follow the setting used in ~\cite{metaconfidence,LST}, using 30(50) samples per class to form the unlabeled set in the 1-shot(5-shot) tasks. For experiments with no distractors in the unlabeled set, the unlabeled set uses the same set of classes as the support set. For experiments with distractors, 3 more distractor classes are added to the unlabeled set. Here we present the results in a semi-supervised setting with 95\% confidence intervals in Table \ref{table:sota_ssl_mini}. It shows that the proposed \ucn{} outperforms all other methods on nearly all benchmarks, indicating the effectiveness of the proposed \ucn{} on semi-supervised tasks.

\begin{figure}[t]
		\centering
		\includegraphics[width=0.4\textwidth]{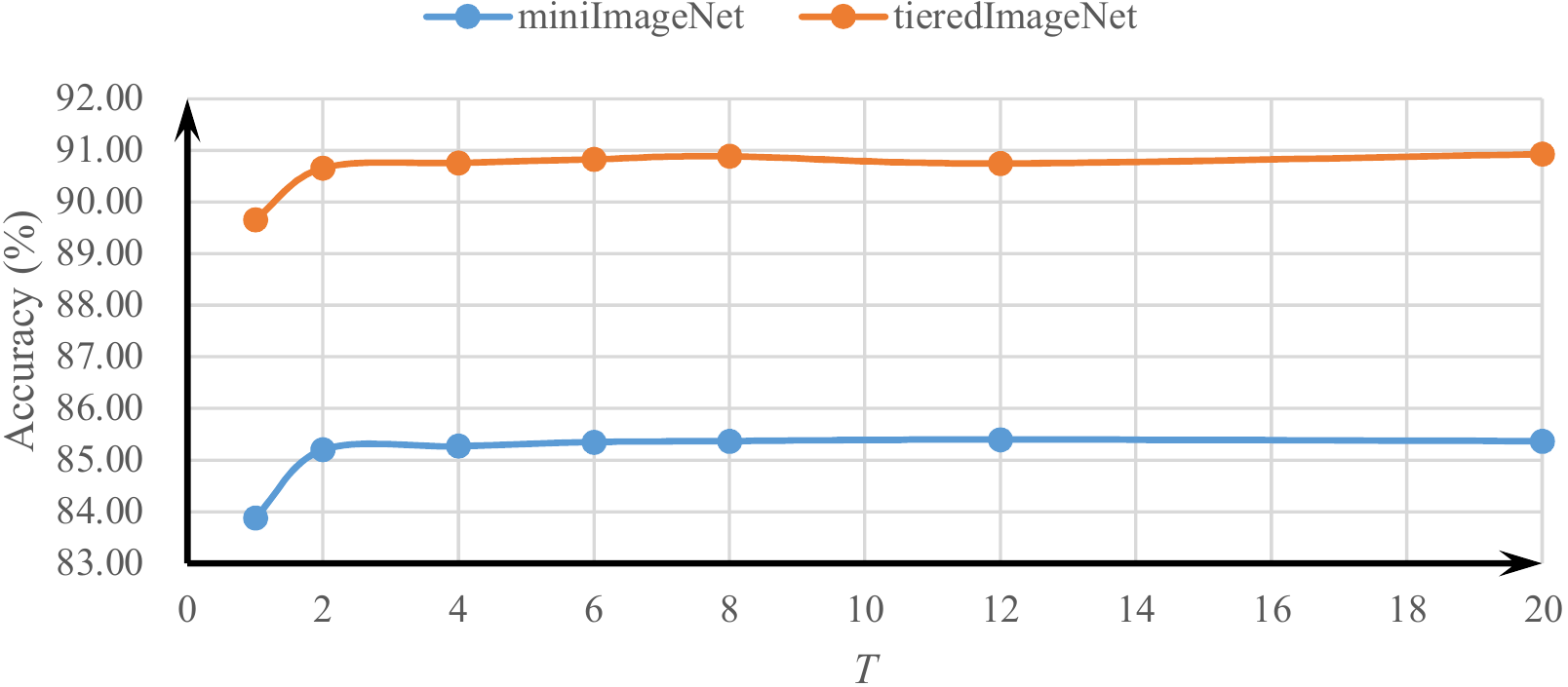}
\caption{Classification accuracy on 5-way 5-shot tasks when different $T$ is adopted.}
\vspace{-0.2in}
\label{fig:abl_T}
\end{figure}
\section{More Ablation Experiments}
\label{smsec:more_abl}

    
\begin{table*}[htbp]
    \begin{center}
    \caption{Details about the augmentation methods used.}
    \label{table:hyp_aug}
    \resizebox{!}{52mm}{ 
    \begin{tabular}{c|>{\arraybackslash}m{0.4\linewidth}|>{\centering\arraybackslash}m{0.2\linewidth}}
    \hline
    
    \hline
    Data Augmentation Algorithms & \multicolumn{1}{c|}{Descriptions} & Hyperparameters \\
    \hline
    
    \hline
    Horizontal flip (F) & Flip the image along the x-axis. & None \\
    \hline
    Histogram equalize (Eq) & Histogram equalization across all the pixels. & None \\
    \hline
    Random crop (X) & Crop an image sized $(H, W)$ to a patch sized $(\lfloor aH \rfloor, \lfloor aW \rfloor)$, and then resize the patch to $(H, W)$. The position of the cropped patch is randomly chosen. & $\frac{1}{\min (H,W)} \leq a \leq 1$ \\
    \hline
    Color adjustment (Co) & Adjust the image color balance with an enhancement factor of $a$. An enhancement factor of $0$ returns a black and white image, and an enhancement factor of $1$ returns the original image. & $a \in \mathbb{R}^*$ \\
    \hline
    Contrast adjustment (Cr) & Adjust the image contrast with an enhancement factor of $a$. An enhancement factor of $0$ returns a solid gray image, and an enhancement factor of $1$ returns the original image. & $a \in \mathbb{R}^*$ \\
    \hline
    Invert (I) & Perform bit-wise inversion on each pixel of the original image $\bm{x}$ to get an inverted image $\bm{x}_{\text{inv}}$, and return the linear combination of both, \ie{}, $\alpha \bm{x} + (1-\alpha) \bm{x}_{\text{inv}}$. & $0 \leq \alpha \leq 1$ \\
    \hline
    Cutout (Cu) & Randomly choose an area sized $(\lfloor aH \rfloor, \lfloor aW \rfloor)$ on a image sized $(H, W)$, and force the value of all pixels in this area to be $127$ (8-bit grayscale) or $(127,127,127)$ (8-bit RGB). & $\frac{1}{\min (H,W)} \leq a \leq 1$ \\
    \hline
    Gaussian blur (B) & Filter an image $\bm{x}$ with  Gaussian blur filter (radius $r$) to get a new image $\bm{x}_{\text{blur}}$, and return the linear combination of both, \ie{}, $\alpha \bm{x} + (1-\alpha) \bm{x}_{\text{blur}}$. & \makecell*[c]{$r \in \mathbb{Z}^+$ \\ $0 \leq \alpha \leq 1$} & &\\
    \hline
    Brightness adjustment (Br) & Adjust the image brightness with an enhancement factor of $a$. An enhancement factor of $0$ returns a black images, and an enhancement factor $1$ returns the original image. & $a \in \mathbb{R}^*$ \\
    \hline
    \hline
    \end{tabular}
    }
    \vspace{-0.2in}
    \end{center}
\end{table*}

    
    

\begin{figure*}[ht]
		\centering
		\includegraphics[width=\textwidth]{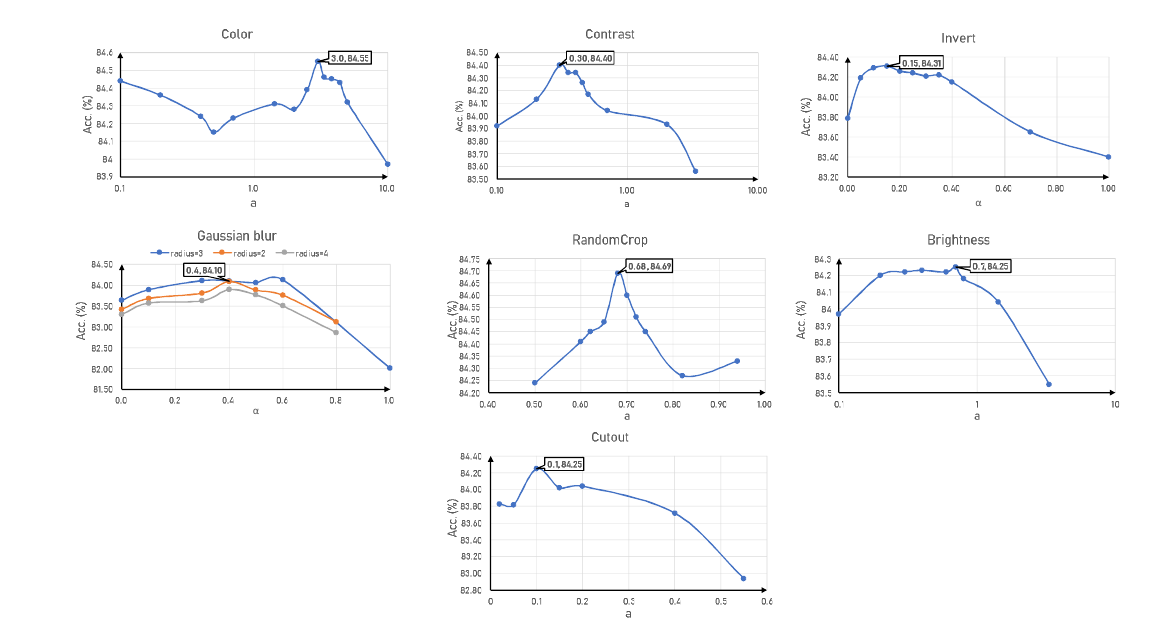}
\caption{Ablation studies on the hyperparameters of data augmentation algorithms. The results are obtained by running experiments on 5-way 5-shot transductive few-shot classification tasks with our UCN, using only the corresponding augmentation in the query set.}
\label{fig:abl_aug_hypparam}
\end{figure*}

\subsection{Hyperparameters of Data Augmentations}
\label{smssec:hypparam_aug}
We will first list the hyperparameters introduced by the data augmentation algorithms involved in our method in Table~\ref{table:hyp_aug}. Then we conduct experiments on 5-way 5-shot transductive few-shot classification tasks with \miniimagenet{} as the dataset to decide the best hyperparameter for certain data augmentation. We will only use \textbf{one} augmentation method in a single experiment. The results are shown in Figure \ref{fig:abl_aug_hypparam} and the finally adopted hyperparameters are marked on the figure.

\subsection{Weight Generator}
\label{smssec:abl_h}
\textbf{Choice of Weights Generator.}
We analyze how the structure of the weights generator $h_\psi(\cdot)$ affects the performance of \ucn{}. Because the weight generator plays a set-to-set function in the paper, Multilayer perceptron (MLP), Gated Recurrent Unit (GRU)~\cite{gru} and multi-headed attention module (MultiAtt)~\cite{transformer} are chosen as the weight generator in our experiments. The experimental results are shown in Table~\ref{table:weight_generate}. It is found that MultiAtt can get the best result. We think it is because the multi-headed attention mechanism can fully extract the association of each element in the set, which is not available in models such as MLP and GRU.




\begin{table}[t]
	\centering
	\fontsize{8}{11}\selectfont   
    \caption{Effect of weight generator on prediction results on 5-way 5-shot task, using \miniimagenet{} as the dataset.}
    \label{table:weight_generate}
	\begin{tabular}{cc}
		\toprule
		Method & Acc(\%) \\
		\midrule
        MLP         & 83.46$\pm$0.77    \\
        GRU         & 84.37$\pm$0.89    \\
\rowcolor{gray!40} MultiAtt    & \textbf{85.35$\pm$0.61}    \\
\hline
\end{tabular}
\vspace{-0.2in}
\end{table}
\subsection{Effect of $T$.}
\label{sssec:abl_T}
\ucn{} refines the prototypes with query instances through an iterative process. Thus, we conduct a series of experiments on \miniimagenet{} and \tieredimagenet{} with ResNet-12 as the backbone. The results are illustrated in Figure~\ref{fig:abl_T}. It can be seen from the figure that the 5-way 5-shot classification accuracy on both \miniimagenet{} and  \tieredimagenet{} increases rapidly when $T\leq 2$, and continues to increase until $T=6$. On the other hand, as can be easily inferred, the time cost per episode continuously increases at a linear pace. As a result, $T=6$ is adopted in all other experiments in the main literature.


\subsection{The Choice of Data Augmentation Algorithms.}
\label{sssec:experiments_ablation_augmentation}
Data augmentation algorithms affect the performance of our method a lot. Some augmentations are too weak to introduce enough perturbations, so using these augmentations won't be very meaningful. 
We conducted a series of experiments to decide the best set of augmentation algorithms. The search space is constructed by 9 different data augmentation algorithms, including horizontal flip(F), histogram equalization(Eq), color adjustment(Co), contrast adjustment(Cr), brightness adjustment(Br), random crop(X), Gaussian blur(B), pixel-wise invert(I) and Cutout(Cu)~\cite{cutout}. 
Table~\ref{table:abl_aug} shows the performance of our method when different sets of data augmentation algorithms are adopted. The results are obtained by running a 5-way 5-shot task using \miniimagenet{} as the dataset. Note that some augmentation algorithms have hyperparameters, and we also select these values with experiments. 
\begin{table}[t]
	\centering
	\fontsize{8}{11}\selectfont   
    \caption{The performance (classification accuracy) of our \ucn{} on 5-way 5-shot tasks with different augmentations adopted, using \miniimagenet{} and \tieredimagenet{} as the datasets. Unit is percentage.}
    \label{table:abl_aug}
    \resizebox{\linewidth}{!}{
	\begin{tabular}{ccccccccccc}
		\toprule
		\multicolumn{9}{c}{Augmentations}   &   \multirow{2}{*}{\textit{mini}} & \multirow{2}{*}{\textit{tiered}}   \\
        F & Eq & X & Co & Cr & I & Cu & Br & B & \\
		\hline\hline
        \checkmark & & & & & & & & & 84.14 
        & 85.53\\
        \checkmark & \checkmark &&&&&&&& 84.40 & 86.12\\
        \checkmark & \checkmark & \checkmark &&&&&&& 85.04 & 86.58\\
        \rowcolor{gray!40}
        \checkmark & \checkmark & \checkmark & \checkmark
        &&&&&& \textbf{85.35} & \textbf{87.83}\\
        & \checkmark & \checkmark & \checkmark
        &&&&&& 84.60 & 86.20\\
        \checkmark & \checkmark & \checkmark & \checkmark & \checkmark &&&&& 85.02 & 87.23\\
        \checkmark & \checkmark & \checkmark & \checkmark &
        \checkmark & \checkmark &&&& 84.89 & 86.67\\
        \checkmark & \checkmark & \checkmark & \checkmark &
        \checkmark & \checkmark & \checkmark &&&
        84.93 & 86.36\\
        \checkmark & \checkmark & \checkmark & \checkmark &
        \checkmark & \checkmark & \checkmark & \checkmark &
        & 84.73 & 86.33\\
        \checkmark & \checkmark & \checkmark & \checkmark &
        \checkmark & \checkmark & \checkmark & \checkmark &
        \checkmark & 84.58 & 85.78\\
\hline
\end{tabular}}
\end{table}
It can be deduced from the table that F, Eq, Co, and X are beneficial to the performance of our method, while Br, B, I, and Cu are not. By comparing row 4 and row 6 of Table~\ref{table:abl_aug}, we can also see that the influence of Cr on the method is trivial. As a result, we decide to use \{F,Eq,Co,X\} as the set of data augmentation algorithms in all experiments.

\bibliographystyle{IEEEbib}
\bibliography{icme2022template}